\documentclass{article}

\usepackage[preprint, nonatbib]{neurips_2021}
\usepackage[utf8]{inputenc} 
\usepackage[T1]{fontenc}    
\usepackage{hyperref}       
\usepackage{url}            
\usepackage{booktabs}       
\usepackage{amsfonts}       
\usepackage{nicefrac}       
\usepackage{microtype}      
\usepackage{xcolor}         

\usepackage{amsmath}

\usepackage{multicol}        
\usepackage{multirow}
\usepackage{changepage}
\usepackage{makecell}
\usepackage{bbding}
\newcommand{\tabincell}[2]{\begin{tabular}{@{}#1@{}}#2\end{tabular}}
\usepackage{amssymb}
\usepackage{pifont}
\usepackage{array}
\newcolumntype{P}[1]{>{\centering\arraybackslash}p{#1}}
\usepackage{siunitx}
\usepackage{bbm}
\usepackage{dsfont}
\usepackage{caption} 
\usepackage{setspace}
\usepackage{balance}
\usepackage{bm} 
\usepackage{color}
\usepackage{stfloats}
\usepackage{graphicx}
\usepackage{subfig}
 \usepackage{mathrsfs}
\usepackage{booktabs} 
\newcommand*\samethanks[1][\value{footnote}]{\footnotemark[#1]}
\title{GODIVA: Generating Open-DomaIn Videos\\ from nAtural Descriptions}
\author{Chenfei Wu$^{1}$\thanks{Both authors contributed equally to this research.} \quad Lun Huang$^{2}$\samethanks[1] \quad Qianxi Zhang$^{1}$ \quad Binyang Li$^{1}$ \\
\textbf{ Lei Ji$^{1}$ \quad Fan Yang$^{1}$ \quad Guillermo Sapiro$^{2}$ \quad Nan Duan$^{1}$\thanks{Corresponding author.} }\\
 {\small $^{1}$Microsoft Research Asia \quad $^{2}$Duke University} \\
{\tt\small\{chewu, qianxi.zhang, binyang.li, leiji, fanyang, nanduan\}@microsoft.com}\\
{\tt\small\{lun.huang, guillermo.sapiro\}@duke.edu}}

\begin{document}

\maketitle

\begin{abstract}
Generating videos from text is a challenging task due to its high computational requirements for training and infinite possible answers for evaluation. Existing works typically experiment on simple or small datasets, where the generalization ability is quite limited. In this work, we propose GODIVA, an open-domain text-to-video pretrained model that can generate videos from text in an auto-regressive manner using a three-dimensional sparse attention mechanism. We pretrain our model on Howto100M, a large-scale text-video dataset that contains more than 136 million text-video pairs. Experiments show that GODIVA not only can be fine-tuned on downstream video generation tasks, but also has a good zero-shot capability on unseen texts. We also propose a new metric called Relative Matching (RM) to automatically evaluate the video generation quality. Several challenges are listed and discussed as future work.
\end{abstract}

\section{Introduction}
``Creativity is a fundamental feature of human intelligence, and a challenge for AI.''~\cite{bodenCreativityArtificialIntelligence1998}. Recent advances in image and text generation have shown great creativity of machines, including GANs~\cite{brockLargeScaleGAN2019,zhangSelfattentionGenerativeAdversarial2019}, VAEs~\cite{razaviGeneratingDiverseHighfidelity2019,oordNeuralDiscreteRepresentation2017}, RNNs\cite{wenSemanticallyConditionedLstmbased2015,prakashNeuralParaphraseGeneration2016} and Self-Attentions~\cite{radfordLanguageModelsAre2019}. However, it is still a challenge for the AI agent to create videos, especially for real-world diversity ones. Generating videos requires the machine to not only create a large number of pixels but also ensure semantic coherence among them. 

\begin{figure}[h]
	\centering
	\includegraphics[width=3.3in]{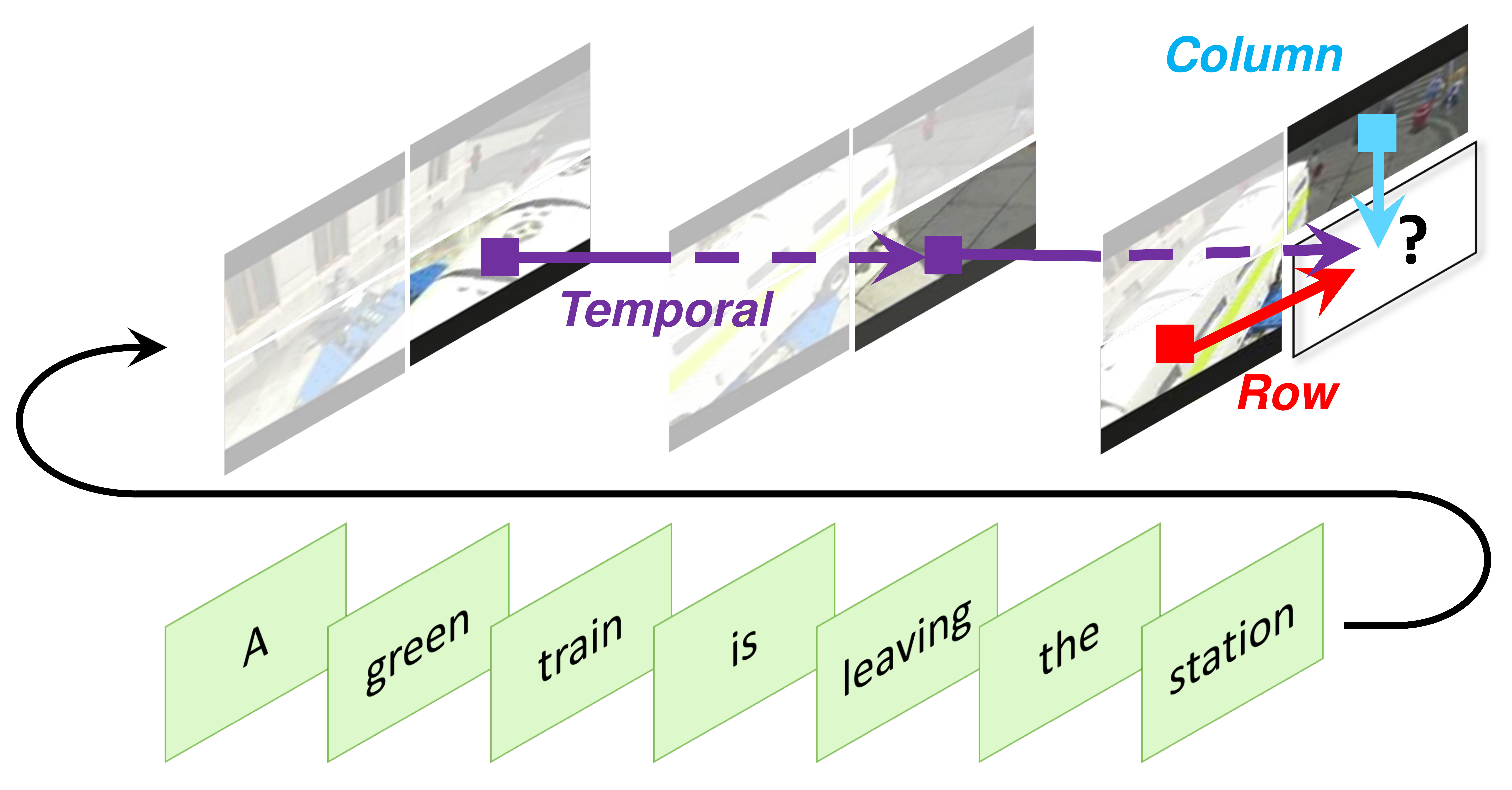}
	\caption{A simple illustration of our GODIVA model with three-dimensional sparse attention mechanism for text-to-video generation task. The video is auto-regressively predicted with the consideration of four aspects: the input text, same position of the previous generated frames, same rows on the same frame, same columns on the same column.}
	\label{fig:small}
\end{figure}

We take up these challenges of generating videos from the text, namely the text-to-video generation (T2V) task. Given a natural description, T2V requires the machine to understand it and create a semantically consistent video. Although not much, there are still some works studying this topic using GANs. Firstly,~\cite{liVideoGenerationText2018} and~\cite{panCreateWhatYou2017} use GANs with 3D convolutions to generate fixed-length low-resolution videos. Then,~\cite{balajiConditionalGANDiscriminative2019} uses a conditional filter to generate videos of varying lengths.~\cite{dengIRCGANIntrospectiveRecurrent2019} integrates LSTM cells with 2D convolutional networks to model both frame quality and temporal coherence. However, these works conduct experiments on simple or small datasets, where generalization ability is limited. 

Besides GAN-based methods, VQ-VAE is another promising research direction and has been shown great progress in generating images and videos, especially DALL-E~\cite{rameshZeroShotTexttoImageGeneration2021} for text-to-image generation. It successfully generates high-quality images from text. In this paper, we turn to the more challenging text-to-video generation task, where both spatial and temporal coherence of the visual information must be taken into account. Some other recent works~\cite{rakhimovLatentVideoTransformer2020,walkerPredictingVideoVQVAE2021,zhangVideoGenGenerativeModeling2020} apply VQ-VAE for the task of video prediction—forecasting future video frames given the past. Concurrently, we are the first to design a VQ-VAE pretrained model for the T2V task.

In this paper, we propose GODIVA to generate open-domain videos from text using VQ-VAE and three-dimensional sparse attention. Firstly, a VQ-VAE auto-encoder is trained to represent continuous video pixels with discrete video tokens. Then, a three-dimensional sparse attention model is trained using language as input and the discrete video tokens as labels to generate videos, considering temporal, column, and row information, as shown in Fig.~\ref{fig:small}.

Our contributions are three-fold: (1) We proposed an open-domain text-to-video pretrained model with a three-dimensional sparse attention mechanism, which can significantly reduce the computation cost; (2) We proposed a new Relative Matching (RM) Metric, which can evaluate both visual quality and semantic match for video generation; (3)We pretrained our proposed model on the HowTo100M dataset and demonstrated its video generation capabilities on both fine-tuning and zero-shot settings.

\section{Related Works}

In this section, we briefly review related works for video generation. We first review the video-to-video generation task, which has been widely studied in recent years. Then we review text-to-image and text-to-video generation. We also highlight the differences between previous models and ours.

\subsection{Video-to-video generation}

Most video generation studies focus on video prediction tasks. Input the first few frames of a video, the video prediction task predicts the following frames of a video. We call it video-to-video (V2V) generation for comparison with text-to-video (T2V) generation. 

Existing video-to-video generation can be divided into three categories. Firstly, deterministic methods directly model the tractable density using RNNs and CNNs and exploit both spatial and temporal information of a video.~\cite{finnUnsupervisedLearningPhysical2016} used ConvLSTM as the basic block to predict pixel motions instead of values.~\cite{lotterDeepPredictiveCoding2017} proposed PredNet, which predicts future frames by incorporating previous predictions. Further,~\cite{wangPredRNNRecurrentNeural2017} proposed stacked ConvLSTM, which shares the hidden state among the layers in the stack. Recently,~\cite{byeonContextvpFullyContextaware2018} proposed ContextVP, which aggregates contextual information for each pixel in all possible directions. Secondly, the GAN-based methods avoid explicit density function and use a generator to generate videos and a discriminator to judge if the video is generated. ~\cite{vondrickGeneratingVideosScene2016} proposed VGAN, which is the first model to generate videos using GANs. After that,~\cite{saitoTemporalGenerativeAdversarial2017} proposed TGAN, which separates a spatiotemporal generator into time-series and space models to generate videos. Then,~\cite{tulyakovMocoganDecomposingMotion2018} proposed MoCoGAN, which produces videos more efficiently by decomposing the latent space into the motion and the content subspaces. Recently,~\cite{saitoTrainSparselyGenerate2020a} proposed TGAN2, which trains each sub-generator with its specific discriminator.  Thirdly, VAE methods model the approximate density by capturing a low-dimensional representation z and optimize a lower bound on the likelihood.~\cite{babaeizadehStochasticVariationalVideo2017} proposed SV2P to capture sequence uncertainty in a single set of latent variables kept fixed for each predicted sequence. Then,~\cite{dentonStochasticVideoGeneration2018} proposed SVG. They used a per-step latent variable (SVG-FP) and a variant with a learned prior (SVG-LP), which makes the prior at a certain timestep a function of previous frames. Recently,~\cite{rakhimovLatentVideoTransformer2020} proposed a Latent Video Transformer, which encodes each frame of a video and predicts the discrete video features. GAN-based models. 

Our model can be categorized into the VAE-based models. Different from recent VQ-VAE based works such as Latent Video Transformer~\cite{rakhimovLatentVideoTransformer2020}, our work focus on  text-to-video generation task instead of video-to-video generation task. We also incorporate a three-dimensional sparse attention to model the sparse relations between visual tokens.

\subsection{Text-to-image generation} \label{sec:texttoimage}
Text-to-image generation has been widely researched in recent years \cite{qiaoMirrorGANLearningTexttoimage2019}. The most similar work is DALL-E~\cite{rameshZeroShotTexttoImageGeneration2021} which successfully generates high-quality images from text. In this paper, we turn to a more challenging text-to-video generation task, which considers both spatial and temporal coherence of the visual information.

\subsection{Text-to-video generation} \label{sec:texttovideo}
Different from video-to-video generation, text-to-video generation has been few studied. Firstly,~\cite{liVideoGenerationText2018} and~\cite{panCreateWhatYou2017} use GANs with 3D convolutions to generate fixed-length low-resolution videos. Then,~\cite{balajiConditionalGANDiscriminative2019} uses a conditional filter to generate videos of varying lengths.~\cite{dengIRCGANIntrospectiveRecurrent2019} integrates LSTM cells with 2D convolutional networks to model both frame quality and temporal coherence.

Most text-to-video generation methods use GAN-based methods while our model incorporates a VQ-VAE for this task. As far as we know, this is the first paper that uses VQ-VAE for this task.

\section{The GODIVA Method}
Let $x$ be an observable video, and we use a discrete latent code $z$ to represent it, which has a lower dimension. In the following, we show how to represent $x$ using $z$ with VQ-VAE~\cite{oordNeuralDiscreteRepresentation2017} in section~\ref{sec:fva}, and generate videos from the text by modeling $P(z|t)$ in section~\ref{sec:GODIVA}, where $t$ denotes the given text.

\begin{figure*}[t]
	\centering
	\includegraphics[width=5.5in]{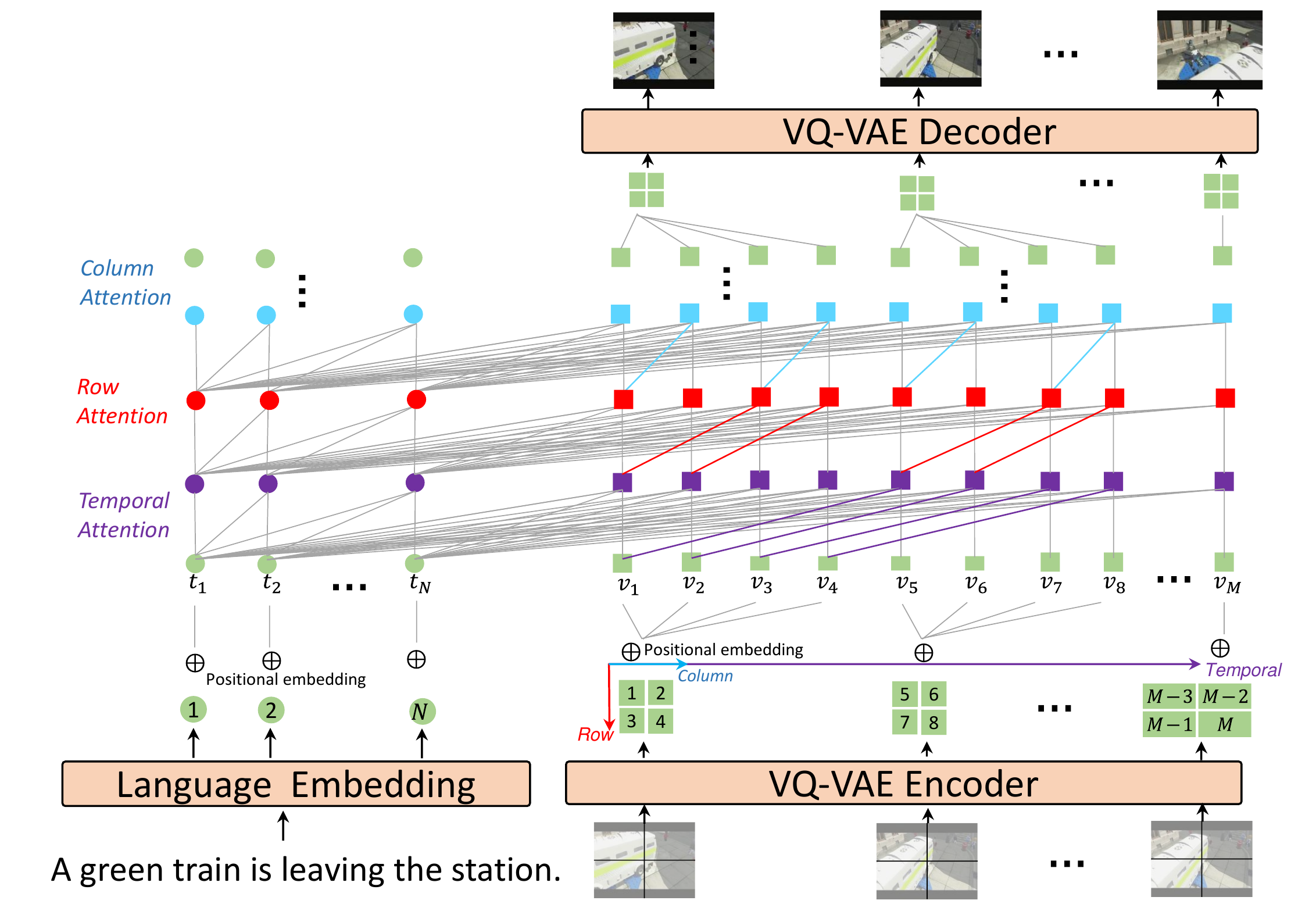}
	\caption{Illustration of GODIVA. To generate a video of $W\times H=64\times 64$ pixels and  $L=10$ frames, the size of the VQ-VAE discrete representation is $w\times h=16\times 16$. Thus the model needs to generate a total of $M=2560$ tokens. When generating the $8^{th}$ visual token, our model only pays attention to the same position token in the previous frame ($4^{th}$ visual token) or the previous row or column token in the same frame ($7^{th}$ and $6^{th}$ visual token).}
	\label{fig:big}
\end{figure*}


\subsection{Frame-wise video auto-encoder} \label{sec:fva}
For an input video $x\in \mathbb{R}^{L\times H\times W\times C}$ with $L$ frames, the $l$th frame $x^{(l)}$ is encoded in Eq.~(\ref{eq:yl}).

\begin{equation} \label{eq:yl} 
y^{(l)}=E(x^{(l)}), \\
\end{equation}
where $y^{(l)}\in \mathbb{R}^{(hw)\times d_B}$ is the latent variable with $h\times w$ regions.Then, $y^{(l)}$ is quantized to get a more compact latent representation, as denoted in Eq.~(\ref{eq:zl}).
\begin{equation} \label{eq:zl} 
z^{(l)}_i=\mathop{\arg\min}_{j}||y_i^{(l)}-B_j||^2,\\
\end{equation}
where $B\in \mathbb{R}^{K\times D}$ is the codebook where the $i$th region of the latent variable $y^{(l)}_i\in \mathbb{R}^{d_B}$ is searched to find the nearest indexes $z^{(l)}\in \mathbb{R}^{hw}$. Then,$z^{(l)}$ is embedded by the codebook in Eq.~(\ref{eq:bl}).

\begin{equation} \label{eq:bl} 
b^{(l)}=B[z^{(l)}],
\end{equation}
where  $b^{(l)}\in \mathbb{R}^{(hw)\times d_B}$ is the embedding of $z^{(l)}$. Next, $b^{(l)}$ is sent to a decoder that reconstructs the original video frame, as shown in Eq.~(\ref{eq:xl2}).
\begin{equation} \label{eq:xl2} 
\hat{x}^{(l)}=D(b^{(l)}),
\end{equation}
where $\hat{x}^{(l)}\in \mathbb{R}^{H\times W\times C}$ is the reconstructed frame. Finally, the  VQ-VAE can be trained in the objective denoted in Eq.~(\ref{eq:lossvae}).
\begin{equation} \label{eq:lossvae} 
\mathscr{L}^{VQ-VAE}=\frac{1}{L}\sum_{l=1}^{L}||x^{(l)}-\hat{x}^{(l)}||_2^2+||sg[y^{(l)}]-b^{(l)}||_2^2+\beta||y^{(l)}-sg[b^{(l)}]||_2^2,
\end{equation}
where the three items are reconstruction loss, codebook loss and commitment loss resepectively. $\beta$ is the weighting factor.$sg$ denotes the stop gradient operator.

\subsection{GODIVA video generator} \label{sec:GODIVA}

In this section, we focus on generating videos from text by modeling the conditional probability $P(z|t)$.
Given an input text $t\in \mathbb{R}^N$ with $N$ tokens, the embeddings of the text are calculated with the consideration of positional information, as denoted in Eq.~(\ref{eq:xe}):
\begin{equation} \label{eq:xe} 
t^e=E^{(t)}[word_{idx}] + P^{(t)}[0,1,...,N-1],
\end{equation}
where $E^{(t)}\in \mathbb{R}^{S\times D}$ is the text embedding matrix, $S$ is the text vocabulary size, $P^{(t)}\in \mathbb{R}^{N\times D}$ is the text positional embedding matrix, and $t^e\in \mathbb{R}^{N\times D}$ is the final text embedding. We use the pretrained VQ-VAE Encoder (Eq.~(\ref{eq:yl}$\sim$\ref{eq:bl})) to encode each frame the ground-truth videos in  Eq.~(\ref{eq:z}):

\begin{equation} \label{eq:z} 
b^{(l)} = B\left[\mathop{\arg\min}_{j}||E(x^{(l)})_j-B_j||^2\right]
\end{equation}
where the ground-truth video sequence $x\in \mathbb{R}^{L\times H\times W\times C}$   is encoded into a sequence of discrete latent visual token embeddings  $b\in \mathbb{R}^{M\times d_B}$. $M=L\times h\times w$ is the maximum of visual tokens. Then, the video embeddings are calculated with the consideration of positional information in Eq.~(\ref{eq:ve}):
\begin{equation} \label{eq:ve} 
v^e=Linear(b) + P^{(v)}[0,1,...,M-1],
\end{equation}
where the Linear layer mapped $z$ to $Linear(z)\in \mathbb{R}^{M\times D}$ , which has the same dimension as $t^e$. $P^{(v)}\in \mathbb{R}^{M\times D}$ is the video positional embedding matrix. $v^e\in \mathbb{R}^{M\times D}$ is the final ground-truth video embedding.  Now, a decoder can be trained to generate videos in an auto-regressive way, as denoted in Eq.~(\ref{eq:vem}):
\begin{equation} \label{eq:vem} 
v^e_{m}=Decoder(t^e, v^e_{<m})
\end{equation}
where $v^e_m\in \mathbb{R}^D$ is the transformed visual embeddings at step $m$. Note that $M$ is a large number, especially for real-word videos. To reduce computation, We introduce a three-dimensional sparse attention in Eq.~(\ref{eq:3d}):
\begin{equation} \label{eq:3d} 
\begin{split}
h_{i,j,l}^{(T)}=SA^{(T)}(v^e_{i,j,<l}), \\
h_{i,j,l}^{(R)}=SA^{(R)}(v^e_{<i,j,l}), \\
h_{i,j,l}^{(C)}=SA^{(C)}(v^e_{i,<j,l}).
\end{split}
\end{equation}
where $SA$ denotes the self-attention layer.$T, R, C$ denotes Temporal, Row, and Column respectively. $h_{i,j,l}^{(T)},h_{i,j,l}^{(R)},h_{i,j,l}^{(C)}\in \mathbb{R}^D$ are the hidden states at step $(i, j, l)$. Note that we change the notation of the step from $m$ to $(i, j, l)$ for a clearer expression of these three sparse attentions. As we can see from Eq.~(\ref{eq:3d}), the sparse attention for each axis only attends to the indexes in the previous axis, instead of the indexes in the global axis.  Thus the computation complexity reduces from $O((Lhw)^2)$ to $O(Lhw(L+h+w))$. Then, the three attention layers are stacked alternately, as denoted in Eq.~(\ref{eq:hijl}).
\begin{equation} \label{eq:hijl} 
h_{ijl}=\underbrace{\left[SA^{(T)},SA^{(R)},SA^{(C)},SA^{(T)},...,SA^{(C)}\right]}_{R~layers}(h_{<=i,<=j,<=l}),
\end{equation}
where $h\in \mathbb{R}^{M\times D}$ is the output hidden states of these stacked attention layers. Then, $h$ is fed to a Linear layer to get the logits of the predicted visual tokens, as denoted in Eq.~(\ref{eq:zhat}).
\begin{equation} \label{eq:zhat} 
P(\hat{z}|t)=\textrm{softmax}(Linear(h))
\end{equation}where the linear layer  maps the dimension of  $h$ into the VQ-VAE vocabulary size $Linear(h) \in \mathbb{R}^{M\times K}$ .$\hat{z} \in \mathbb{R}^{M}$ is the predicted visual tokens. Finally, the model is trained using cross-entropy loss, as denoted in Eq.~(\ref{eq:l2}).
\begin{equation} \label{eq:l2} 
\mathscr{L}=-\frac{1}{M}\sum_{i=1}^{M}z_ilog(P(\hat{z}|t))
\end{equation}

\section{Experiments}

\subsection{Datasets}
We pretrain GODIVA on Howto100M dataset~\cite{miechHowto100mLearningTextvideo2019}, which consists of more than 136 million text-video pairs. We then evaluate our model on the MSR-VTT dataset~\cite{xuMsrvttLargeVideo2016}, which consists of 10000 video clips with 20 human-annotated captions for each of them. We also train GODIVA from scratch on the Moving Mnist dataset~\cite{mittalSyncdrawAutomaticVideo2017} and Double Moving Mnist dataset, both were automatically generated from the Mnist dataset~\cite{LeCun_Backpropagationappliedhandwritten_1989}. The original Moving Mnist dataset has two motions: up-down and left-right. In this paper, we follow~\cite{dengIRCGANIntrospectiveRecurrent2019} and add four more directions: move left then right, move right then left, move up then down and move down then up.
\subsection{Evaluation Metrics}
It is challenging to quantitatively evaluate the performance of text-to-video generation models. This is mainly due to two reasons: Firstly, given a piece of text, there are countless corresponding videos. It is hard to objectively judge which one is better. Secondly, an evaluation metric should consider both visual quality and semantic matching of the generated video. To handle these challenges, we introduce two kind of metrics: A CLIP Similarity (SIM) metric and a Relative Matching (RM) metric for automatic evaluation in Sec.~\ref{sec:rmm},  A Visual Realisticity (VR) and Semantic Consistency (SC) metric for human evaluation in Sec.~\ref{sec:msm}.

\subsubsection{Automatic Evaluation Metrics} \label{sec:rmm}
The key factor for judging the quality of the generated video is whether it matches the text. Using a pretrained visual-language matching model will inevitably introduce the bias of its domain data. Thanks to recent zero-shot work CLIP~\cite{radfordLearningTransferableVisual2021}, which provides a strong zero-shot ability for visual-text matching and thus reduced those data biases. Since CLIP is pretrained between image and text, we calculate the similarities between text and each frame of the video and then take the average value as the semantic matching in Eq.~(\ref{eq:sim}).

   \begin{equation} \label{eq:sim} 
SIM(t, \hat{v})= \frac{1}{L}\sum_{l=1}^{L}CLIP(t, \hat{v}^{(l)}),
\end{equation}where $t$ denotes the input text. $\hat{v}$ is the predicted video with $L$ frames.  Note that $SIM$ only provides the absolute score of the semantic match. To further reduce the influence of the CLIP model, we divide $SIM$ by the similarity between text and the ground-truth video to get a relative matching score, which we call Relative Matching (RM) metric, as denoted in Eq.~(\ref{eq:rm}).
\begin{equation} \label{eq:rm} 
RM(t, \hat{v})= \frac{SIM(t, \hat{v})}{SIM(t, v)},
\end{equation}
where $v$ is the ground-truth video with $L$ frames. The RM metric reveals the domain-independent generation quality since if the generated video is more relevant to the text, it will obviously have a higher RM value. If the generated video is not relevant to the text or has a low quality, the RM value will be lower.

\subsubsection{Human Evaluation Metrics} \label{sec:msm}
To conduct a human evaluation, we invite 200 evaluators as testees and conduct a human evaluation.  Let $\{M_1, M_2,.., M_N\}$ be a set of models to evaluate, $T$ be the number of samples in the test set. To reduce the subjective biases, we ask the testees to compare the Visual Realisticity ($VR$) and Semantic Consistency ($SC$) of two videos ($v_i, v_j$) generated from two models ($M_i, M_j$) with the same query $q$ respectively, as denoted in Eq.~(\ref{eq:vrijt})$\sim$(\ref{eq:scijt}).
\begin{equation} \label{eq:vrijt} 
VR(M_i)=\frac{1}{NT}\sum^{N,T} r_{ij}^{(t)},~~~r_{ij}^{(t)}=\left\{
\begin{aligned}
1 & , & v_i~\text{is~more~realistic~than~}v_j~\text{for sample}~t, \\
0 & , & otherwise.
\end{aligned}
\right.
\end{equation}
\begin{equation} \label{eq:scijt} 
 SC(M_i)=\frac{1}{NT}\sum^{N,T} c_{ij}^{(t)},~~~c_{ij}^{(t)}=\left\{
\begin{aligned}
1 & , & v_i~\text{is~more~consistent~with~} q \text{~than~}v_j~\text{for sample}~t, \\
0 & , & otherwise.
\end{aligned}
\right.
\end{equation}

\newcommand{\visuwidth}{1.6cm}

\begin{figure*}[h]
  \centering
  \begin{tabular}{lccccc}
\specialrule{0.12em}{0pt}{2pt}
    \tabincell{l}{\emph{Play golf on grass.} \\ T2V~\cite{liVideoGenerationText2018} \\ (64$\times$ 64)} & \raisebox{-.5\height}{\includegraphics[width=\visuwidth ]{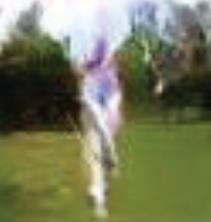}} & \raisebox{-.5\height}{\includegraphics[width=\visuwidth ]{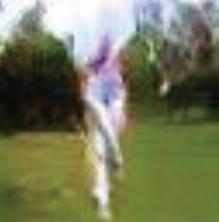}} & \raisebox{-.5\height}{\includegraphics[width=\visuwidth]{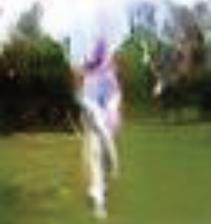}} & \raisebox{-.5\height}{\includegraphics[width=\visuwidth]{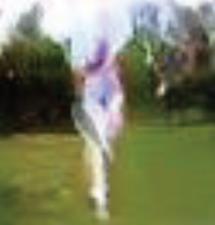}} & \raisebox{-.5\height}{\includegraphics[width=\visuwidth ]{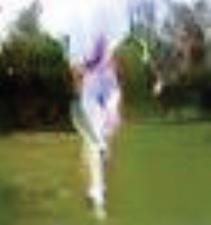}}  \\\specialrule{0.05em}{1.1pt}{1.1pt}
    \tabincell{l}{\emph{Play golf on grass.} \\ TFGAN~\cite{balajiConditionalGANDiscriminative2019}\\ (128$\times$ 128)} & \raisebox{-.5\height}{\includegraphics[width=\visuwidth ]{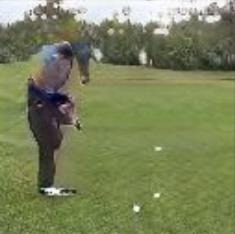}} & \raisebox{-.5\height}{\includegraphics[width=\visuwidth ]{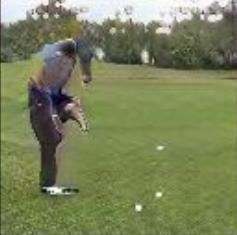}} & \raisebox{-.5\height}{\includegraphics[width=\visuwidth ]{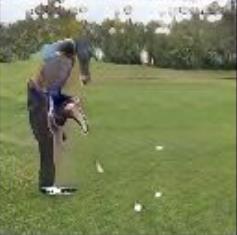}} & \raisebox{-.5\height}{\includegraphics[width=\visuwidth ]{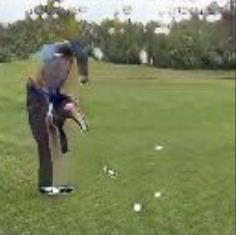}} & \raisebox{-.5\height}{\includegraphics[width=\visuwidth ]{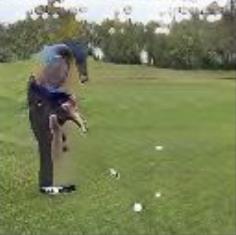}} \\\specialrule{0.05em}{1.1pt}{1.1pt}
    \tabincell{l}{\emph{Play golf on grass.} \\ GODIVA(ours) \\ (64$\times$64)} & \raisebox{-.5\height}{\includegraphics[width=\visuwidth ]{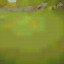}} & \raisebox{-.5\height}{\includegraphics[width=\visuwidth ]{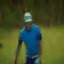}} & \raisebox{-.5\height}{\includegraphics[width=\visuwidth ]{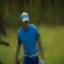}} & \raisebox{-.5\height}{\includegraphics[width=\visuwidth ]{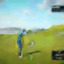}} & \raisebox{-.5\height}{\includegraphics[width=\visuwidth ]{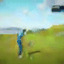}}  \\\specialrule{0.05em}{1.1pt}{1.1pt}
    \tabincell{l}{\emph{Play golf on grass.} \\GODIVA(ours) \\ (128$\times$128)} & \raisebox{-.5\height}{\includegraphics[width=\visuwidth ]{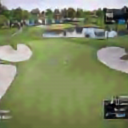}} & \raisebox{-.5\height}{\includegraphics[width=\visuwidth ]{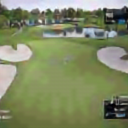}} & \raisebox{-.5\height}{\includegraphics[width=\visuwidth ]{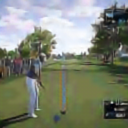}} & \raisebox{-.5\height}{\includegraphics[width=\visuwidth ]{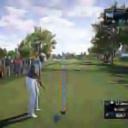}} & \raisebox{-.5\height}{\includegraphics[width=\visuwidth ]{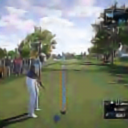}}  \\\specialrule{0.05em}{1.1pt}{1.1pt}
    \tabincell{l}{\emph{A baseball game is played.} \\ GODIVA(ours) \\ (128$\times$128)} & \raisebox{-.5\height}{\includegraphics[width=\visuwidth ]{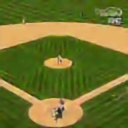}} & \raisebox{-.5\height}{\includegraphics[width=\visuwidth ]{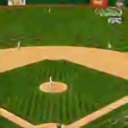}} & \raisebox{-.5\height}{\includegraphics[width=\visuwidth]{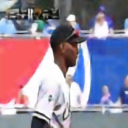}} & \raisebox{-.5\height}{\includegraphics[width=\visuwidth ]{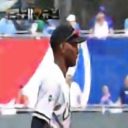}} & \raisebox{-.5\height}{\includegraphics[width=\visuwidth]{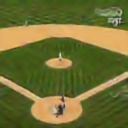}}  \\
 \specialrule{0.12em}{2pt}{1pt}
  \end{tabular}
  \caption{Comparison of samples from our model to prior approaches on real-word dataset. n$\times$n represents the number of pixels of the video frame.}
  \label{fig:cmp}
\end{figure*}

\begin{figure*}[h]
  \centering
  \begin{tabular}{p{2cm}p{5.6cm}p{5.6cm}}
\specialrule{0.12em}{0pt}{2pt}
    Model & \tabincell{l}{Input Sentence: Digit 9 is moving down\\ then up.} &  \tabincell{l}{Input Sentence: Digit 7 moves right then \\ left while digit 3 moves down then up.}\\\specialrule{0.05em}{1.1pt}{1.1pt}
    VGAN\cite{vondrickGeneratingVideosScene2016} & \raisebox{-.5\height}{\includegraphics[width=\linewidth ]{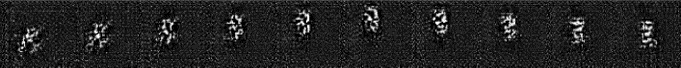}} & \raisebox{-.5\height}{\includegraphics[width=\linewidth ]{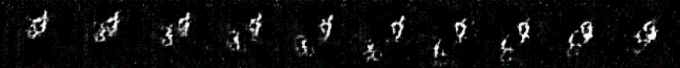}} \\\specialrule{0em}{1.1pt}{1.1pt}
    SyncDraw\cite{mittalSyncdrawAutomaticVideo2017} & \raisebox{-.5\height}{\includegraphics[width=\linewidth ]{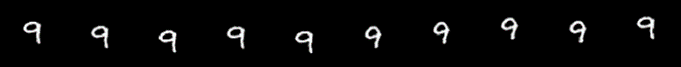}} & \raisebox{-.5\height}{\includegraphics[width=\linewidth ]{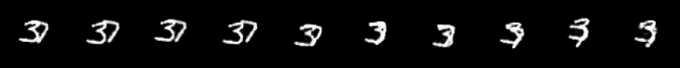}} \\\specialrule{0em}{1.1pt}{1.1pt}
    TGANs\cite{panCreateWhatYou2017} & \raisebox{-.5\height}{\includegraphics[width=\linewidth ]{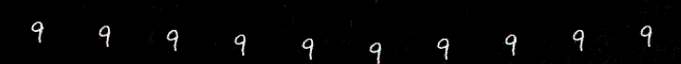}} & \raisebox{-.5\height}{\includegraphics[width=\linewidth ]{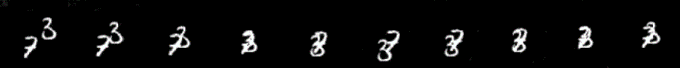}} \\\specialrule{0em}{1.1pt}{1.1pt}
    MocoGAN\cite{tulyakovMocoganDecomposingMotion2018} & \raisebox{-.5\height}{\includegraphics[width=\linewidth ]{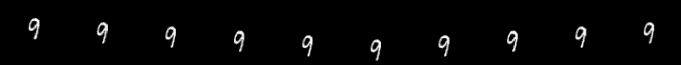}} & \raisebox{-.5\height}{\includegraphics[width=\linewidth ]{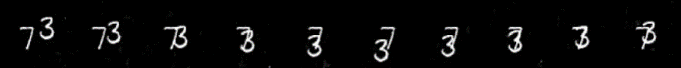}} \\\specialrule{0em}{1.1pt}{1.1pt}
    IRC-GAN\cite{dengIRCGANIntrospectiveRecurrent2019} & \raisebox{-.5\height}{\includegraphics[width=\linewidth ]{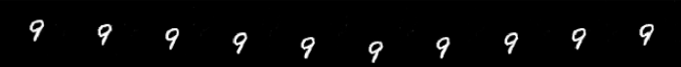}} & \raisebox{-.5\height}{\includegraphics[width=\linewidth ]{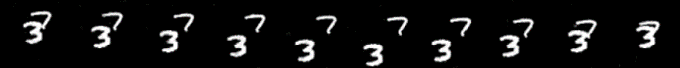}} \\\specialrule{0em}{1.1pt}{1.1pt}
    GODIVA(ours) & \raisebox{-.5\height}{\includegraphics[width=\linewidth ]{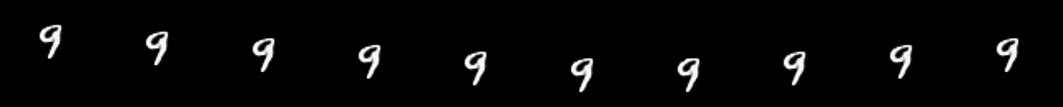}} & \raisebox{-.5\height}{\includegraphics[width=\linewidth ]{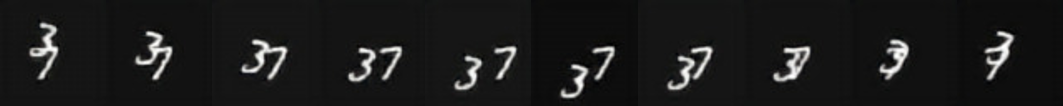}} \\\specialrule{0em}{1.1pt}{1.1pt}
 \specialrule{0.12em}{2pt}{1pt}
  \end{tabular}
  \caption{Comparison of samples from our model to prior approaches on Moving Mnist and Double Moving Mnist dataset. Note that VGAN, TGANs and MocoGAN are a modifed version by \cite{dengIRCGANIntrospectiveRecurrent2019} to support text-to-video generation.}
  \label{fig:moving}
\end{figure*}

\subsection{Implementation details}
In Sec.~\ref{sec:fva}, the size of the input video is $L=10, H=64, W=64, C=3$. Both the encoder $E$ in Eq.~(\ref{eq:yl}) and the decoder $D$ in Eq.~(\ref{eq:xl2}) are implemented with two CNN layers. The kernel size is 4 and stride is 2. Thus the latent variable has the size of $h\times w=16\times 16$. The latent variable dimension $d_B=128$. The VQ-VAE codebook has a total of $K=10000$ tokens. The VQ-VAE model is pretrained on ImageNet with a learning rate of 1e-3 and batch size 32. Note that when we conduct experiments on Moving Mnist Dataset, we train another VQ-VAE on this dataset. We found this will lead to better generation performance.

In Sec.~\ref{sec:GODIVA}, the input text has a maximum length of $N=35$. The dimension $D=1024$. The maximum size of the visual tokens is $M=2560$. The Self-Att in Eq.~(\ref{eq:3d}) uses 16 attention heads. The GODIVA has a total of $R=12$ layers in Eq.~(\ref{eq:hijl}). The GODIVA model is pretrained on the Howto100M dataset with 64 V100 GPUs. It is finetuned on the MSR-VTT dataset with 8 V100 GPUs. Both settings have the same batch size of 32 and a learning rate of 5e-4. More details, including the source code, will come soon in Github.

\begin{figure}%
    \centering
    \subfloat[\centering SIM scores of text and ground-truth videos]{{\includegraphics[width=5.2cm]{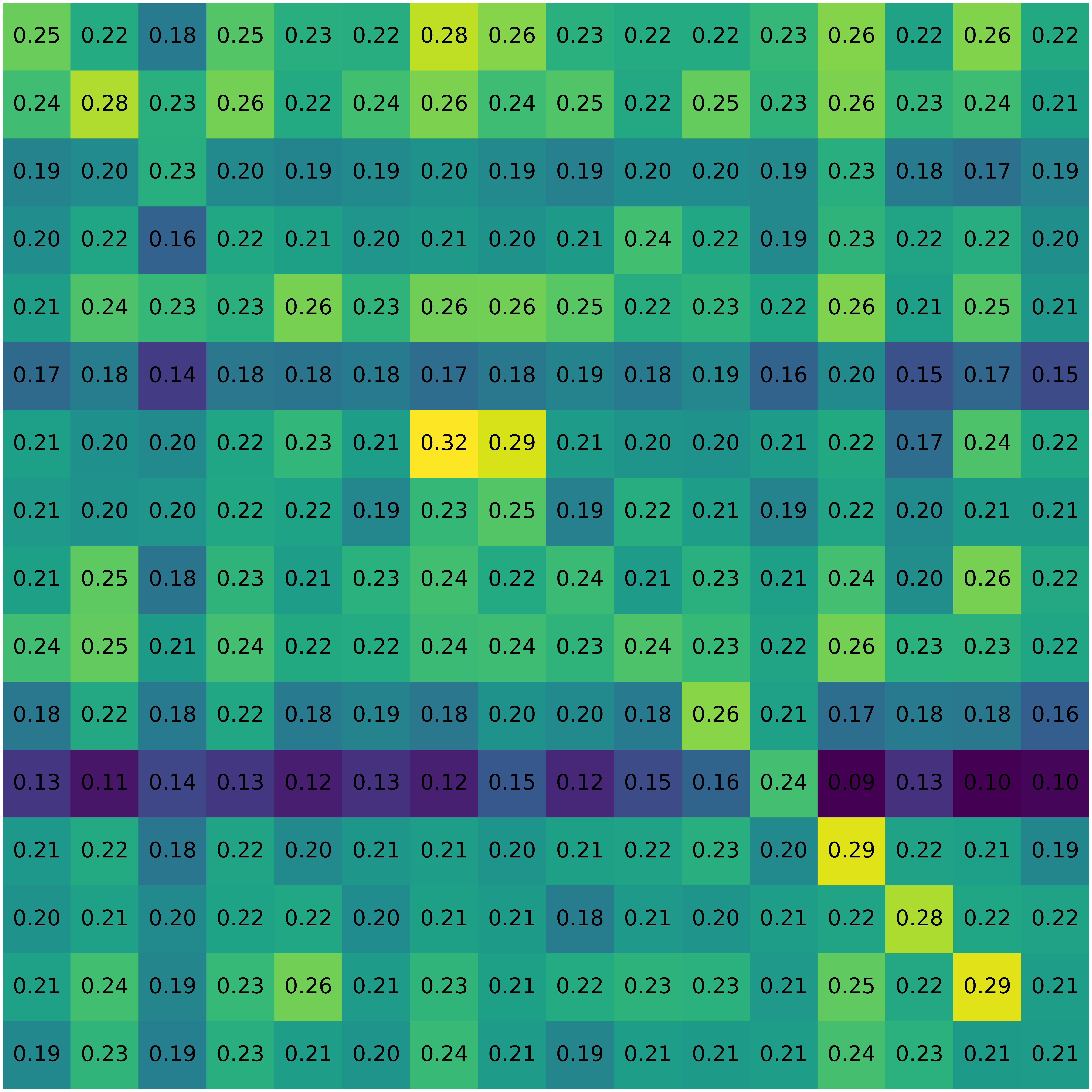} }}%
    \qquad
    \subfloat[\centering SIM scores of text and predicted videos by GODIVA w/ CLIP ranking]{{\includegraphics[width=5.2cm]{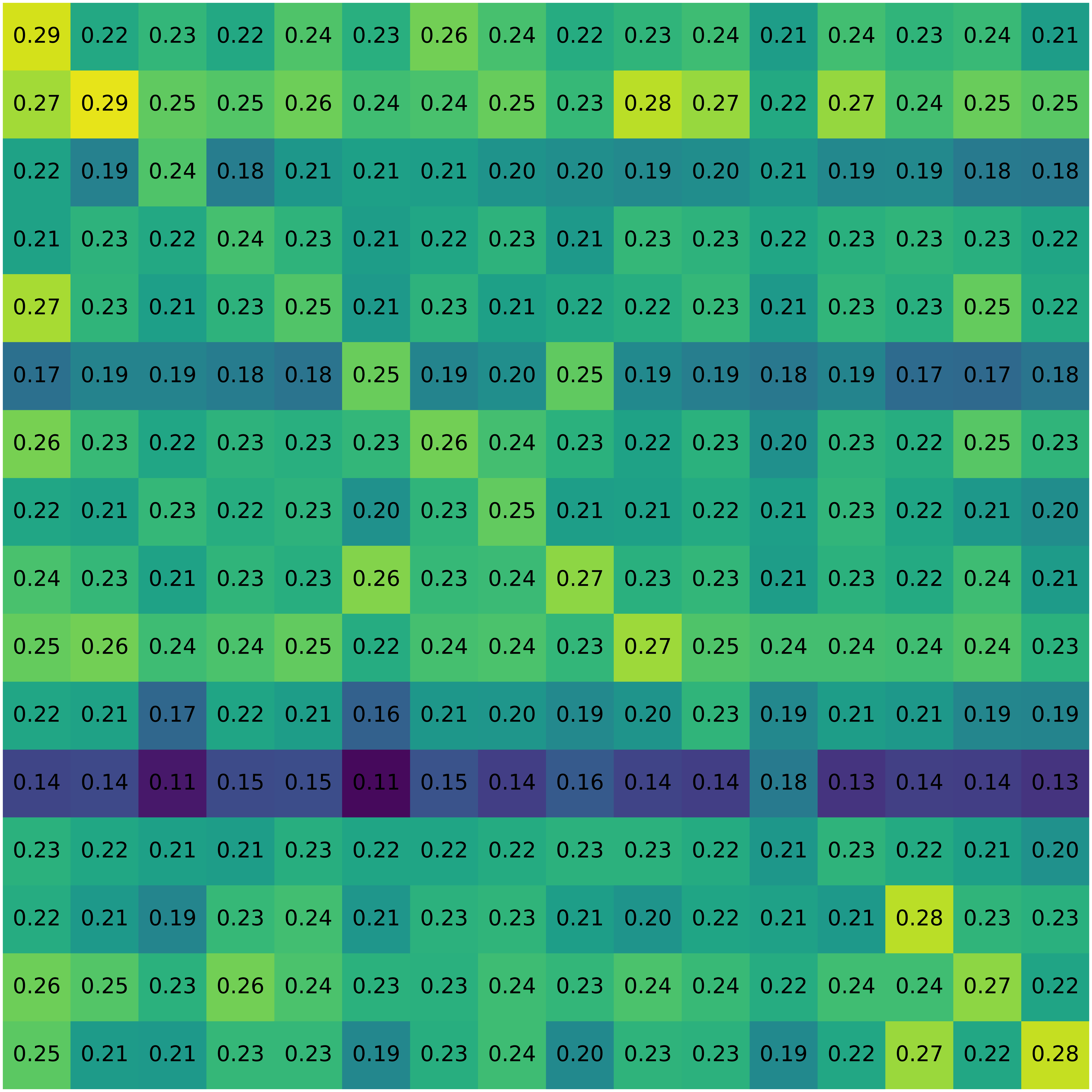} }}%
    \caption{SIM scores calculated in 16 random samples. In sub figure (a) and (b), row denotes the query and the column denotes the videos.}%
    \label{fig:matrix}%
\end{figure}

\newcommand{\pwidth}{1.3cm}
\begin{table*}[h]
\begin{center}
\caption{Qualitative Results on MSR-VTT dataset. All values reported
are multiplied by 100.}
\label{tab:mod}
\begin{tabular}{p{6cm}p{\pwidth}p{\pwidth}p{\pwidth}p{\pwidth}}
\toprule
\multicolumn{1}{c}{\multirow{2}{*}{Model}}&\multicolumn{2}{c}{Automatic Evaluation} & \multicolumn{2}{c}{Human Evaluation}\\
\cmidrule{2-3} \cmidrule{4-5}
 & SIM & RM & VR & SC  \\
\midrule
GT & 24.23 & 100 & - & -   \\

GODIVA (6 layer) & 21.45 & 86.94 & 9.38 & 9.38 \\
GODIVA w/o Row Attention & 21.23 & 85.95 & 31.25 & 40.63   \\
GODIVA w/o Temporal Attention  & 21.52 & 87.44 & 40.63 & 43.75  \\
GODIVA w/o Column Attention  & 22.08 & 89.20 & 46.88 & 46.88  \\
GODIVA  & 22.82 & 93.48 & 81.25 & 78.13\\
GODIVA w/ CLIP ranking  & 24.02 & 98.34 & 88.12 & 81.25  \\
\bottomrule
\end{tabular}
\label{tab:metric}
\end{center}
\end{table*}

\subsection{Qualitative Results}

We qualitatively evaluate our model from two aspects. 
Firstly, we evaluate the zero-shot ability of our model by comparing GODIVA to two prior approaches: T2V~\cite{liVideoGenerationText2018} and TFGAN~\cite{balajiConditionalGANDiscriminative2019}. Both approaches are trained on the real-world dataset created from a clean-up of  Kinetics~\cite{kayKineticsHumanAction2017} and Youtube videos. As shown in Fig.~(\ref{fig:cmp}), for the same query "Play golf on grass", T2V successfully generates the grass and action of "playing golf" in a resolution of 64$\times$64, but the result looks blurry (see the first row). TFGAN was successfully trained on a resolution of 128$\times$128 and generates a higher quality result (see the second row).  Both T2V and TFGAN generate text-related videos, but the generated frames are in a single scene and the difference between frames is not significant. This limits the creativity of neural models. Interestingly, GODIVA not only generates text-related videos but also changing scenes (see the third and fourth row). For example, GODIVA(64$\times$64) first shows the grass field, then it gives the athlete a close-up shot, and finally the action of hitting the golf ball. Note that GODIVA(64$\times$64) and GODIVA(128$\times$128) are different models, thus they generate totally different videos.  The last row gives another (128$\times$128) resolution results generated by GODIVA. In total, GODIVA is able to generate videos with clear frames and coherent semantics.

Secondly, we evaluate the unseen video generation ability by comparing GODIVA to several GAN-based approaches. The models in Fig.~(\ref{fig:moving}) are trained on the Moving MNIST dataset~\cite{mittalSyncdrawAutomaticVideo2017} and Double Moving MNIST dataset~\cite{mittalSyncdrawAutomaticVideo2017} respectively. Note that there is no video in the training set that shows "Digit 9 is moving down than up", but there are some variant samples such as "Digit 9 is moving left and right" or "Digit 3 is moving down than up". GODIVA successfully generates semantic correct results(see the last row of the left part). This shows GODIVA learns to capture the semantic alignment between text and video, rather than just search videos in the training set to find the one most similar to the input sentence. Besides, GODIVA generates high-quality videos, even compared with the state-of-the-art IRC-GAN~\cite{mittalSyncdrawAutomaticVideo2017} approach. The digit "9" is both spatially clear and temporally consistent. Another example in Double Moving MNIST on the right shows a similar phenomenon.

\subsection{Quantitative Results}
We quantitatively evaluate our model through both automatic and human metrics. To validate the effectiveness of RM metric, we first draw the SIM scores between text and ground-truth videos in Fig.~\ref{fig:matrix}(a). It can be seen from the diagonal that SIM is basically able to distinguish semantically similar videos from other videos. Tab.~\ref{tab:metric} shows the ablation experiments of different settings of GODIVA. We pretrain GODIVA on Howto100M dataset and finetune it on MSR-VTT dataset.We find that SIM and RM have the same trend with the human evaluation metrics. The first row shows the results between input text and ground-truth videos. The second row shows that sufficient scale for GODIVA is crucial. GODIVA (6 layer) shows worse performance than the default GODIVA setting (12 layer). The next three rows show the effectiveness of the three dimensional attentions. We found that the Row Attention is the most important. Following DALL-E~\cite{rameshZeroShotTexttoImageGeneration2021}, we randomly sample 32 times in the top 10 probabilities in Eq.~(\ref{eq:zhat}) during inference and using CLIP ranking to find the best generated video. The performance is then significantly improved to 98.34 in RM metric.

\section{Conclusions}
In this paper, we propose a three-dimensional sparse attention to generate open-domain videos from natural descriptions using VQ-VAE discrete visual tokens. We also propose a new Relative Matching metric to automatically evaluate generation quality. Experiments show that our model not only can be fine-tuned on downstream video generation tasks, but also has a good zero-shot capability on unseen texts. However, there are still several challenges:
Firstly, it is still a great challenge to generate long videos with high resolution. When generating only $64\times 64$ resolution videos with 10 frames, the total of visual tokens $M$ already becomes 2560. Secondly, automatically evaluating text-to-video generation task remains a challenge. In the future, video-based CLIP metric may give more accurate results for semantic consistency for text and videos. Thirdly, GAN-based methods show a great potential for text-to-video generation (see Fig.~(\ref{fig:moving})), their generative abilities for open-domain dataset remain a good research direction.
\bibliographystyle{plain}
\bibliography{ref}

\section{Appendix}

\newcommand{\awidth}{13cm}

\begin{figure*}[h]
  \centering
  \begin{tabular}{cl}
\specialrule{0.12em}{0pt}{2pt}

\multicolumn{2}{l}{Input Text: A man with suit sitting on the chair talking in front of the camera.} \\
GT:&\raisebox{-.5\height}{\includegraphics[width=\awidth]{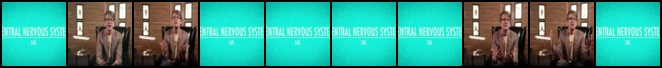}}  \\
Pred:&\raisebox{-.5\height}{\includegraphics[width=\awidth]{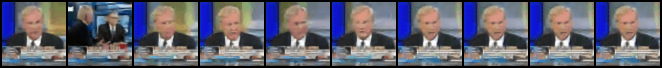}}  \\\specialrule{0.05em}{2pt}{1.1pt}

\multicolumn{2}{l}{Input Text: A ballroom dance class.} \\
GT:&\raisebox{-.5\height}{\includegraphics[width=\awidth]{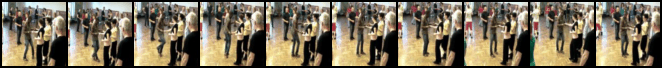}}  \\
Pred:&\raisebox{-.5\height}{\includegraphics[width=\awidth]{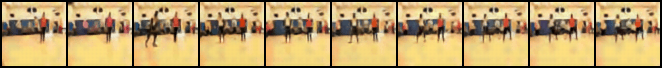}}  \\\specialrule{0.05em}{2pt}{1.1pt}
\multicolumn{2}{l}{Input Text: A band preforms a song.} \\
GT:&\raisebox{-.5\height}{\includegraphics[width=\awidth]{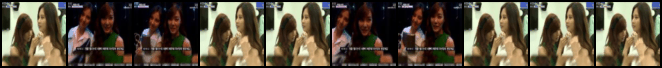}}  \\
Pred:&\raisebox{-.5\height}{\includegraphics[width=\awidth]{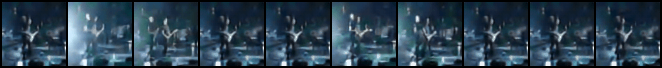}}  \\\specialrule{0.05em}{2pt}{1.1pt}
\multicolumn{2}{l}{Input Text: A baseball batter hits the ball to the fence and a outfielder goes after it.} \\
GT:&\raisebox{-.5\height}{\includegraphics[width=\awidth]{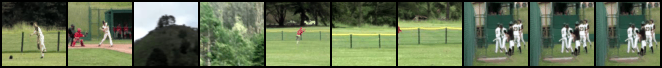}}  \\
Pred:&\raisebox{-.5\height}{\includegraphics[width=\awidth]{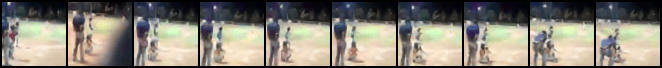}}  \\\specialrule{0.05em}{2pt}{1.1pt}

\multicolumn{2}{l}{Input Text: A person is preparing some art.} \\
GT:&\raisebox{-.5\height}{\includegraphics[width=\awidth]{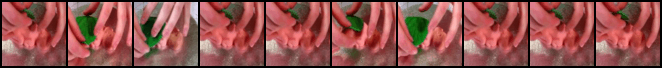}}  \\
Pred:&\raisebox{-.5\height}{\includegraphics[width=\awidth]{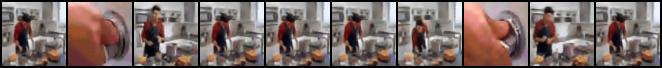}}  \\\specialrule{0.05em}{2pt}{1.1pt}

\multicolumn{2}{l}{Input Text: A gamer talks about his minecraft experience.} \\
GT:&\raisebox{-.5\height}{\includegraphics[width=\awidth]{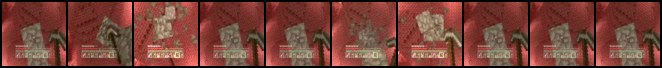}}  \\
Pred:&\raisebox{-.5\height}{\includegraphics[width=\awidth]{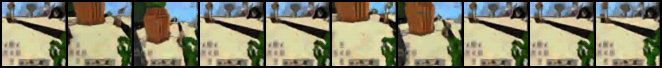}}  \\\specialrule{0.05em}{2pt}{1.1pt}
\multicolumn{2}{l}{Input Text: A girl and someone is putting an painted egg in to a water.} \\
GT:&\raisebox{-.5\height}{\includegraphics[width=\awidth]{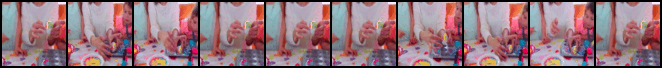}}  \\
Pred:&\raisebox{-.5\height}{\includegraphics[width=\awidth]{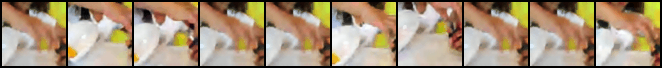}}  \\\specialrule{0.05em}{2pt}{1.1pt}

\specialrule{0.12em}{2pt}{1pt}
 
  \end{tabular}
  \caption{More samples generated by GODIVIA.}
\end{figure*}

\begin{figure*}[h]
  \centering
  \begin{tabular}{cl}
\specialrule{0.12em}{0pt}{2pt}

\multicolumn{2}{l}{Input Text: A girl and the judges talking on the the voice.} \\
GT:&\raisebox{-.5\height}{\includegraphics[width=\awidth]{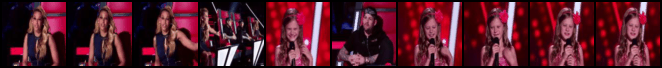}}  \\
Pred:&\raisebox{-.5\height}{\includegraphics[width=\awidth]{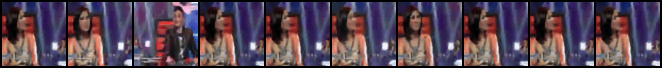}}  \\\specialrule{0.05em}{2pt}{1.1pt}
\multicolumn{2}{l}{Input Text: A family doing exercises indoors.} \\
GT:&\raisebox{-.5\height}{\includegraphics[width=\awidth]{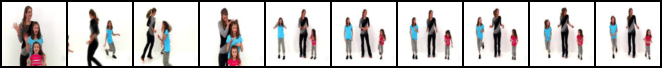}}  \\
Pred:&\raisebox{-.5\height}{\includegraphics[width=\awidth]{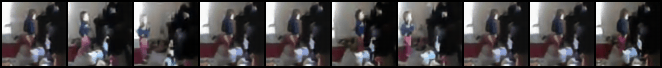}}  \\\specialrule{0.05em}{2pt}{1.1pt}

\multicolumn{2}{l}{Input Text: The girl puts foundation to her cute face with a sponge.} \\
GT:&\raisebox{-.5\height}{\includegraphics[width=\awidth]{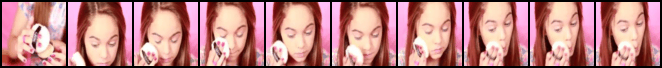}}  \\
Pred:&\raisebox{-.5\height}{\includegraphics[width=\awidth]{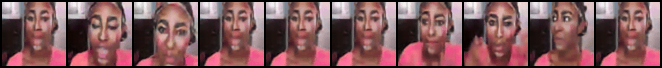}}  \\\specialrule{0.05em}{2pt}{1.1pt}

\multicolumn{2}{l}{Input Text: A guy talks about the features of the jeep cherokee.} \\
GT:&\raisebox{-.5\height}{\includegraphics[width=\awidth]{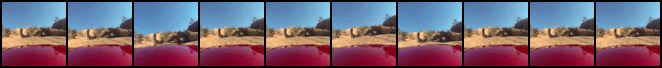}}  \\
Pred:&\raisebox{-.5\height}{\includegraphics[width=\awidth]{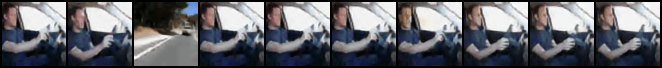}}  \\\specialrule{0.05em}{2pt}{1.1pt}
\multicolumn{2}{l}{Input Text: A guy working on his engine with multiple parts.} \\
GT:&\raisebox{-.5\height}{\includegraphics[width=\awidth]{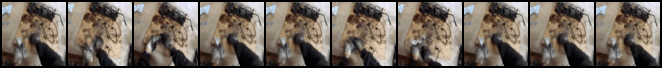}}  \\
Pred:&\raisebox{-.5\height}{\includegraphics[width=\awidth]{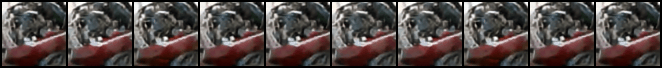}}  \\\specialrule{0.05em}{2pt}{1.1pt}

\multicolumn{2}{l}{Input Text: There is a brown hair woman walking on the ramp.} \\
GT:&\raisebox{-.5\height}{\includegraphics[width=\awidth]{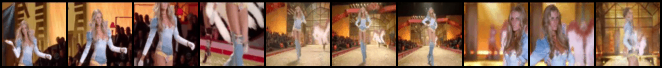}}  \\
Pred:&\raisebox{-.5\height}{\includegraphics[width=\awidth]{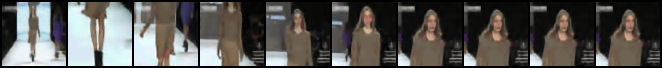}}  \\\specialrule{0.05em}{2pt}{1.1pt}

\multicolumn{2}{l}{Input Text: In a kitchenthere is women preparing some fried dishes.} \\
GT:&\raisebox{-.5\height}{\includegraphics[width=\awidth]{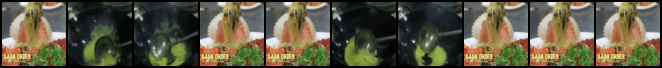}}  \\
Pred:&\raisebox{-.5\height}{\includegraphics[width=\awidth]{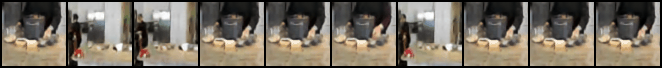}}  \\\specialrule{0.05em}{2pt}{1.1pt}

  \end{tabular}
  \caption{More samples generated by GODIVIA.}
\end{figure*}

\end{document}